# Toward human-centered shared autonomy AI paradigms for human-robot teaming in healthcare

**Reza Abiri**, **Ali Rabiee**, **Sima Ghafoori**, **Anna Cetera**

*Translational Neurorobotics Laboratory (https://web.uri.edu/tnlab/)*
Dept. of Electrical, Computer and Biomedical Engineering
University of Rhode Island
Kingston, RI, USA

Contact Information: reza_abiri@uri.edu
*All authors contributed equally to this white paper report.*

## 1. Abstract

With recent advancements in AI and computation tools, intelligent paradigms emerged to empower different fields such as healthcare robots with new capabilities. Advanced AI robotic algorithms (e.g., reinforcement learning) can be trained and developed to autonomously make individual decisions to achieve a desired and usually fixed goal. However, such independent decisions and goal achievements might not be ideal for a healthcare robot that usually interacts with a dynamic end-user or a patient. In such a complex human-robot interaction (teaming) framework, the dynamic user continuously wants to be involved in decision-making as well as introducing new goals while interacting with their present environment in real-time. To address this challenge, an adaptive shared autonomy AI paradigm is required to be developed for the two interactive agents (Human & AI agents) with a foundation based on human-centered factors to avoid any possible ethical issues and guarantee no harm to humanity.

## 2. Robotics in healthcare and human-robot interaction

The application of the robotics field in healthcare has expanded during the past years, particularly with the spread of COVID-19. As a remotely accessible tool to patients, nursing robots [1] and surgical robots [2] have been studied by different scientific groups. These robots are usually required to be controlled by an operator or a surgeon through a shared control paradigm. Brain-controlled assistive robots, an emerging category in healthcare robotics, have been introduced to facilitate a close physical human-robot interaction framework to restore the physical movements of severely paralyzed patients for independent living [3-12]. Additionally, rehabilitation robots have been used with direct contact with patients (e.g., strokes) to improve motor performance by recovery of cortical plasticity [13-17]. Social robots were designed to assist elderly patients with cognitive tasks [8, 18].

## 3. Traditional shared control paradigms

Developing a blended, intelligent, and balanced framework of human-robot teamwork has become an essential element, particularly in healthcare robots where direct human (operator, patient) engagement plays an important role in overall reported performance satisfaction [19, 20]. To this end, different shared control paradigms were developed to combine human and robot inputs through sequential [8-10], parallel, or blended structures [14]. In most of these studies, traditional and simple control and machine learning algorithms were used to achieve a mutual goal. For example, noninvasive joystick movements [21] or peripheral signals [22-25] were used as a portion of inputs alongside human input into a shared controlled assistive robotic arm platform to complete reach tasks with an inverse kinematics model. Invasive Brain-controlled approaches [3-7] were employed to share the control for a reach task using traditional machine learning techniques. A limited number of studies have targeted the application of shared control theories in restoring reach tasks in stroke patients using robotic exoskeletons [14-16].

## 4. Innovative shared autonomy paradigms

With advancements in computational tools over the past decade, the field of robotics has also been influenced by learning algorithms such as the game theory algorithm [26] and reinforcement learning (RL) [27, 28]. Particularly, complex RL with deep learning layer mechanisms (deep RL) has become popular in learning the representation of an object for autonomous manipulation [29-31], performing dexterous manipulation tasks [29, 32-36], semantics learning of grasp [37] and end-to-end visuomotor control [38, 39] using robotic arms. Leveraging these algorithms for modeling the shared autonomy dynamics in human-robot interaction introduces the possibility of a more intuitive interaction with healthcare robots. In this shared autonomy paradigm [40], rather than using shared control strategies such as sequential inputs generated by the human and robot [41] or a simple summation of those inputs [14], an AI algorithm (deep RL) mechanism can provide an approach/method? to blend the human and robotic inputs for advancing adaptive and intuitive control.

## 5. Human-centered AI for shared autonomy paradigms

In recent years, the concept of human-AI collaborative algorithms was revised to human-centered AI (HCAI) algorithms for human-AI teaming [42, 43] in different fields. Human, technology, and ethics are the three main factors in HCAI paradigms [44, 45]. Each factor is essential for the design of an effective human-AI team. For example, if the final product of a human-AI robotic system only considers AI technology and ethics but ignores human factors (e.g., authority, needs, user experience), the system may become unusable or even harmful to the user [20, 46, 47]. As reported in previous studies, fatigue and cognitive load experienced by the users were parts of translational barriers in brain-controlled high-DOF robotic arms [4, 6, 11, 48] Human-centered AI factors is essential and will be prominent in shared autonomy paradigms where two agents (human, AI agent) interact and collaborate to achieve the human's desired goals. AI agents should always be supervised by the user to guarantee ethical regulation and safe experiences.

## 6. A case study

A critical challenge in developing assistive robotic systems lies in enabling intuitive control and seamless human-robot interaction when the user's input capabilities are severely limited. Many end-users, such as those with impaired motor function or residual muscular control, may have access to only low degrees of freedom (DOF) or low-bandwidth control interfaces. However, sought-after robotic assistance tasks such as dexterous object manipulation often requires high-dimensional control in a three-dimensional space. Mapping these low-DOF user inputs, such as facial gestures, breath control, or limited extremity movements, onto the full 6-DOF control space for robotic manipulation presents a highly non-trivial mapping challenge. Current high-DOF robotic manipulators generally lack the intelligence and flexibility to adaptively address this assistance level tradeoff through appropriate boosting algorithms that can satisfy users with limited motor function. In such complex interactive dynamics, an assistive artificial intelligence (AI) agent is crucial for not only generating the remaining motor function required to complete high-dimensional manipulation tasks but also addressing the desired assistance level tradeoff. The developed shared embodied machine intelligence should understand, boost, and shape the end-users limited function, enabling collaborative completion of dexterous manipulation tasks in an intuitive, natural manner that meets daily needs. To address this gap, we have developed an AI-based algorithm that utilizes computer vision techniques to create a solid framework for human-robot teaming even with low-DOF user input. In our preliminary results, we have successfully amplified a 1-DOF user input to a 3-DOF manipulation task. Figure 1 shows an example of our algorithm implemented on a Kinova Jaco2 assistive robotic arm, where the human provides input across the y-axis (left or right), and the AI agent amplifies this input to all three axes, enabling the robot to complete pick-and-place tasks.

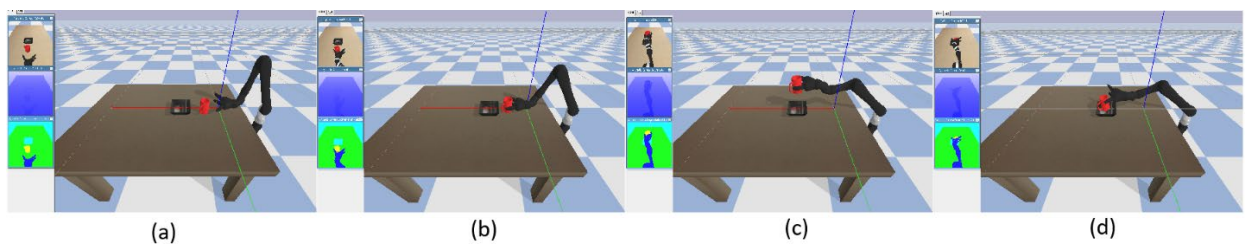

*Figure 1: A successful attempt for a pick-and-place task*

## References

[1] N. Maalouf, A. Sidaoui, I. H. Elhajj, and D. Asmar, "Robotics in nursing: a scoping review," *Journal of Nursing Scholarship,* vol. 50, no. 6, pp. 590-600, 2018.

[2] B. S. Peters, P. R. Armijo, C. Krause, S. A. Choudhury, and D. Oleynikov, "Review of emerging surgical robotic technology," *Surgical endoscopy,* vol. 32, pp. 1636-1655, 2018.

[3] J. L. Collinger, B. Wodlinger, J. E. Downey, W. Wang, E. C. Tyler-Kabara, D. J. Weber, A. J. McMorland, M. Velliste, M. L. Boninger, and A. B. Schwartz, "High-performance neuroprosthetic control by an individual with tetraplegia," *The Lancet,* vol. 381, no. 9866, pp. 557-564, 2013.

[4] J. E. Downey, J. M. Weiss, K. Muelling, A. Venkatraman, J.-S. Valois, M. Hebert, J. A. Bagnell, A. B. Schwartz, and J. L. Collinger, "Blending of brain-machine interface and vision-guided autonomous robotics improves neuroprosthetic arm performance during grasping," *Journal of neuroengineering and rehabilitation,* vol. 13, no. 1, pp. 1-12, 2016.

[5] M. S. Fifer, G. Hotson, B. A. Wester, D. P. McMullen, Y. Wang, M. S. Johannes, K. D. Katyal, J. B. Helder, M. P. Para, and R. J. Vogelstein, "Simultaneous neural control of simple reaching and grasping with the modular prosthetic limb using intracranial EEG," *IEEE transactions on neural systems and rehabilitation engineering,* vol. 22, no. 3, pp. 695-705, 2013.

[6] L. R. Hochberg, D. Bacher, B. Jarosiewicz, N. Y. Masse, J. D. Simeral, J. Vogel, S. Haddadin, J. Liu, S. S. Cash, and P. van der Smagt, "Reach and grasp by people with tetraplegia using a neurally controlled robotic arm," *Nature,* vol. 485, no. 7398, pp. 372-375, 2012.

[7] J. Vogel, S. Haddadin, B. Jarosiewicz, J. Simeral, D. Bacher, L. Hochberg, J. Donoghue, and P. Van Der Smagt, "An assistive decision-and-control architecture for force-sensitive hand–arm systems driven by human–machine interfaces," *The International Journal of Robotics Research,* vol. 34, no. 6, pp. 763-780, 2015.

[8] R. Abiri, J. McBride, X. Zhao, and Y. Jiang, "A real-time brainwave based neuro-feedback system for cognitive enhancement," in *ASME 2015 Dynamic Systems and Control Conference (Columbus, OH)*, 2015.

[9] R. Abiri, X. Zhao, G. Heise, Y. Jiang, and F. Abiri, "Brain computer interface for gesture control of a social robot: An offline study," in *Electrical Engineering (ICEE), 2017 Iranian Conference on*, 2017: IEEE, pp. 113-117.

[10] R. Abiri, J. Kilmarx, M. Raji, and X. Zhao, "Planar Control of a Quadcopter Using a Zero-Training Brain Machine Interface Platform," in *Biomedical Engineering Society Annual Meeting (BMES 2016)*, 2016.

[11] N. Natraj, S. Seko, R. Abiri, H. Yan, Y. Graham, A. Tu-Chan, E. F. Chang, and K. Ganguly, "Flexible regulation of representations on a drifting manifold enables long-term stable complex neuroprosthetic control," *bioRxiv,* 2023.

[12] S. Seko, N. Natraj, R. Abiri, Y. Graham, N. Hardy, E. F. Chang, and K. Ganguly, "ECoG based BCI control of reaching and grasping with a robotic arm using stacked discrete commands," in *Society for Neuroscience Annual Meeting (SfN 2021)*, Chicago, IL, USA, Nov. 08-11 2021.

[13] A. Budhota, K. S. Chua, A. Hussain, S. Kager, A. Cherpin, S. Contu, D. Vishwanath, C. W. Kuah, C. Y. Ng, and L. H. Yam, "Robotic Assisted Upper Limb Training Post Stroke: A Randomized Control Trial Using Combinatory Approach Toward Reducing Workforce Demands," *Frontiers in neurology,* vol. 12, 2021.

[14] A. Sarasola-Sanz, N. Irastorza-Landa, E. López-Larraz, C. Bibián, F. Helmhold, D. Broetz, N. Birbaumer, and A. Ramos-Murguialday, "A hybrid brain-machine interface based on EEG and EMG activity for the motor rehabilitation of stroke patients," in *2017 International conference on rehabilitation robotics (ICORR)*, 2017: IEEE, pp. 895-900.

[15] G. Rosati, J. E. Bobrow, and D. J. Reinkensmeyer, "Compliant control of post-stroke rehabilitation robots: using movement-specific models to improve controller performance," in *ASME International Mechanical Engineering Congress and Exposition*, 2008, vol. 48630, pp. 167-174.

[16] E. T. Wolbrecht, V. Chan, D. J. Reinkensmeyer, and J. E. Bobrow, "Optimizing compliant, model-based robotic assistance to promote neurorehabilitation," *IEEE Transactions on Neural Systems and Rehabilitation Engineering,* vol. 16, no. 3, pp. 286-297, 2008.

[17] S. Ghafoori, Wellington, K., Garneau, B., Koning, A., Broadmeadow, E., Norouzi, M., Jouaneh, M., Shahriari, Y., Abiri, R., "Development of a novel seamless magnetic-based actuated planar robotic platform for upper limb extremity rehabilitation," in *Society for Neuroscience (SfN)*, Washington DC, November 11-15 2023.

[18] R. Abiri, X. Zhao, and Y. Jiang, "A Real Time EEG-Based Neurofeedback platform for Attention Training," in *Biomedical Engineering Society Annual Meeting (BMES 2016)*, 2016.

[19] E. You and K. Hauser, "Assisted teleoperation strategies for aggressively controlling a robot arm with 2d input," in *Robotics: science and systems*, 2012, vol. 7: MIT Press USA, p. 354.

[20] D.-J. Kim, R. Hazlett-Knudsen, H. Culver-Godfrey, G. Rucks, T. Cunningham, D. Portee, J. Bricout, Z. Wang, and A. Behal, "How autonomy impacts performance and satisfaction: Results from a study with spinal cord injured subjects using an assistive robot," *IEEE Transactions on Systems, Man, and Cybernetics-Part A: Systems and Humans,* vol. 42, no. 1, pp. 2-14, 2011.

[21] S. Javdani, H. Admoni, S. Pellegrinelli, S. S. Srinivasa, and J. A. Bagnell, "Shared autonomy via hindsight optimization for teleoperation and teaming," *The International Journal of Robotics Research,* vol. 37, no. 7, pp. 717-742, 2018.

[22] J. Vogel, J. Bayer, and P. Van Der Smagt, "Continuous robot control using surface electromyography of atrophic muscles," in *2013 IEEE/RSJ International Conference on Intelligent Robots and Systems*, 2013: IEEE, pp. 845-850.

[23] J. Vogel and A. Hagengruber, "An sEMG-based interface to give people with severe muscular atrophy control over assistive devices," in *2018 40th Annual International Conference of the IEEE Engineering in Medicine and Biology Society (EMBC)*, 2018: IEEE, pp. 2136-2141.

[24] J. Vogel, A. Hagengruber, M. Iskandar, G. Quere, U. Leipscher, S. Bustamante, A. Dietrich, H. Höppner, D. Leidner, and A. Albu-Schäffer, "EDAN: An EMG-controlled Daily Assistant to Help People With Physical Disabilities," in *2020 IEEE/RSJ International Conference on Intelligent Robots and Systems (IROS)*, 2020: IEEE, pp. 4183-4190.

[25] K. Z. Zhuang, N. Sommer, V. Mendez, S. Aryan, E. Formento, E. D'Anna, F. Artoni, F. Petrini, G. Granata, and G. Cannaviello, "Shared human–robot proportional control of a dexterous myoelectric prosthesis," *Nature Machine Intelligence,* vol. 1, no. 9, pp. 400-411, 2019.

[26] Y. Li, G. Carboni, F. Gonzalez, D. Campolo, and E. Burdet, "Differential game theory for versatile physical human–robot interaction," *Nature Machine Intelligence,* vol. 1, no. 1, pp. 36-43, 2019.

[27] J. Kober, J. A. Bagnell, and J. Peters, "Reinforcement learning in robotics: A survey," *The International Journal of Robotics Research,* vol. 32, no. 11, pp. 1238-1274, 2013.

[28] L. Wang, Y. Xiang, W. Yang, A. Mousavian, and D. Fox, "Goal-auxiliary actor-critic for 6D robotic grasping with point clouds," in *Conference on Robot Learning*, 2022: PMLR, pp. 70-80.

[29] S. Gu, E. Holly, T. Lillicrap, and S. Levine, "Deep reinforcement learning for robotic manipulation with asynchronous off-policy updates," in *2017 IEEE international conference on robotics and automation (ICRA)*, 2017: IEEE, pp. 3389-3396.

[30] S. Joshi, S. Kumra, and F. Sahin, "Robotic Grasping using Deep Reinforcement Learning," *arXiv preprint arXiv:2007.04499,* 2020.

[31] A. Franceschetti, E. Tosello, N. Castaman, and S. Ghidoni, "Robotic Arm Control and Task Training through Deep Reinforcement Learning," *arXiv preprint arXiv:2005.02632,* 2020.

[32] H. Zhu, A. Gupta, A. Rajeswaran, S. Levine, and V. Kumar, "Dexterous manipulation with deep reinforcement learning: Efficient, general, and low-cost," in *2019 International Conference on Robotics and Automation (ICRA)*, 2019: IEEE, pp. 3651-3657.

[33] S. Gu, E. Holly, T. Lillicrap, and S. Levine, "Deep reinforcement learning for robotic manipulation," *arXiv preprint arXiv:1610.00633,* vol. 1, 2016.

[34] S. Joshi, S. Kumra, and F. Sahin, "Robotic grasping using deep reinforcement learning," in *2020 IEEE 16th International Conference on Automation Science and Engineering (CASE)*, 2020: IEEE, pp. 1461-1466.

[35] V. François-Lavet, P. Henderson, R. Islam, M. G. Bellemare, and J. Pineau, "An introduction to deep reinforcement learning," *arXiv preprint arXiv:1811.12560,* 2018.

[36] S. Levine, E. Holly, S. Gu, and T. Lillicrap, "Deep reinforcement learning for robotic manipulation," ed: Google Patents, 2019.

[37] E. Jang, S. Vijayanarasimhan, P. Pastor, J. Ibarz, and S. Levine, "End-to-end learning of semantic grasping," *arXiv preprint arXiv:1707.01932,* 2017.

[38] S. Levine, C. Finn, T. Darrell, and P. Abbeel, "End-to-end training of deep visuomotor policies," *The Journal of Machine Learning Research,* vol. 17, no. 1, pp. 1334-1373, 2016.

[39] Y. Zhu, Z. Wang, J. Merel, A. Rusu, T. Erez, S. Cabi, S. Tunyasuvunakool, J. Kramár, R. Hadsell, and N. de Freitas, "Reinforcement and imitation learning for diverse visuomotor skills," *arXiv preprint arXiv:1802.09564,* 2018.

[40] S. Reddy, A. D. Dragan, and S. Levine, "Shared autonomy via deep reinforcement learning," *arXiv preprint arXiv:1802.01744,* 2018.
[41] C. Xie, Q. Yang, Y. Huang, S. W. Su, T. Xu, and R. Song, "A Hybrid Arm-hand Rehabilitation Robot with EMG-based Admittance Controller," *IEEE Transactions on Biomedical Circuits and Systems,* 2021.
[42] B. Shneiderman, *Human-centered AI*. Oxford University Press, 2022.
[43] O. Ozmen Garibay, B. Winslow, S. Andolina, M. Antona, A. Bodenschatz, C. Coursaris, G. Falco, S. M. Fiore, I. Garibay, and K. Grieman, "Six human-centered artificial intelligence grand challenges," *International Journal of Human–Computer Interaction,* vol. 39, no. 3, pp. 391-437, 2023.
[44] W. Xu and Z. Gao, "Applying human-centered AI in developing effective human-AI teaming: A perspective of human-AI joint cognitive systems," *arXiv preprint arXiv:2307.03913,* 2023.
[45] W. Xu and M. Dainoff, "Enabling human-centered AI: A new junction and shared journey between AI and HCI communities," *Interactions,* vol. 30, no. 1, pp. 42-47, 2023.
[46] S. Doncieux, R. Chatila, S. Straube, and F. Kirchner, "Human-centered AI and robotics," *AI Perspectives,* vol. 4, no. 1, pp. 1-14, 2022.
[47] T. Bhattacharjee, E. K. Gordon, R. Scalise, M. E. Cabrera, A. Caspi, M. Cakmak, and S. S. Srinivasa, "Is more autonomy always better? exploring preferences of users with mobility impairments in robot-assisted feeding," in *Proceedings of the 2020 ACM/IEEE international conference on human-robot interaction*, 2020, pp. 181-190.
[48] C. H. Blabe, V. Gilja, C. A. Chestek, K. V. Shenoy, K. D. Anderson, and J. M. Henderson, "Assessment of brain–machine interfaces from the perspective of people with paralysis," *Journal of neural engineering,* vol. 12, no. 4, p. 043002, 2015.